\documentclass[10pt,twocolumn,letterpaper]{article}

\usepackage{cvpr}
\usepackage{times}
\usepackage{epsfig}
\usepackage{graphicx}
\usepackage{amsmath}
\usepackage{amssymb}
\usepackage{stmaryrd}
\usepackage[misc]{ifsym}
\usepackage[linesnumbered,ruled]{algorithm2e}

\usepackage[x11names,dvipsnames,table]{xcolor} 
\usepackage{multirow}
\usepackage{array}
\newcolumntype{L}[1]{>{\raggedright\let\newline\\\arraybackslash\hspace{0pt}}m{#1}}
\newcolumntype{C}[1]{>{\centering\let\newline\\\arraybackslash\hspace{0pt}}m{#1}}
\newcolumntype{R}[1]{>{\raggedleft\let\newline\\\arraybackslash\hspace{0pt}}m{#1}}

\usepackage[pagebackref=true,breaklinks=true,letterpaper=true,colorlinks,bookmarks=false]{}

 \cvprfinalcopy 

\def\cvprPaperID{1199} 

\ifcvprfinal\fi
\begin{document}

\title{Iterative Reorganization with Weak Spatial Constraints:\\Solving Arbitrary Jigsaw Puzzles for Unsupervised Representation Learning}

\author{Chen Wei\textsuperscript{1}, Lingxi Xie\textsuperscript{2}, Xutong Ren\textsuperscript{1}, Yingda Xia\textsuperscript{2}, Chi Su\textsuperscript{3}, Jiaying Liu\textsuperscript{1}, Qi Tian\textsuperscript{4}, Alan L. Yuille\textsuperscript{2}\\
Peking University\textsuperscript{1} \quad The Johns Hopkins University\textsuperscript{2} \quad Kingsoft\textsuperscript{3} \quad Huawei Noah's Ark Lab\textsuperscript{4}\\
{{\tt\small weichen582@pku.edu.cn}\quad{\tt\small 198808xc@gmail.com}\quad{\tt\small tonghelen@pku.edu.cn}\quad{\tt\small yxia25@jhu.edu}}\\
{{\tt\small suchi@kingsoft.com}\quad{\tt\small liujiaying@pku.edu.cn}\quad{\tt\small tian.qi1@huawei.com}\quad{\tt\small alan.l.yuille@gmail.com}}
}

\maketitle

\begin{abstract}
Learning visual features from unlabeled image data is an important yet challenging task, which is often achieved by training a model on some annotation-free information. We consider spatial contexts, for which we solve so-called jigsaw puzzles, i.e., each image is cut into grids and then disordered, and the goal is to recover the correct configuration. Existing approaches formulated it as a classification task by defining a fixed mapping from a small subset of configurations to a class set, but these approaches ignore the underlying relationship between different configurations and also limit their application to more complex scenarios.

This paper presents a novel approach which applies to jigsaw puzzles with an arbitrary grid size and dimensionality. We provide a fundamental and generalized principle, that weaker cues are easier to be learned in an unsupervised manner and also transfer better. In the context of puzzle recognition, we use an iterative manner which, instead of solving the puzzle all at once, adjusts the order of the patches in each step until convergence. In each step, we combine both unary and binary features on each patch into a cost function judging the correctness of the current configuration. Our approach, by taking similarity between puzzles into consideration, enjoys a more reasonable way of learning visual knowledge. We verify the effectiveness of our approach in two aspects. First, it is able to solve arbitrarily complex puzzles, including high-dimensional puzzles, that prior methods are difficult to handle. Second, it serves as a reliable way of network initialization, which leads to better transfer performance in a few visual recognition tasks including image classification, object detection, and semantic segmentation.
\end{abstract}

\section{Introduction}
\label{Introduction}

\begin{figure}
\centering
\includegraphics[width=8cm]{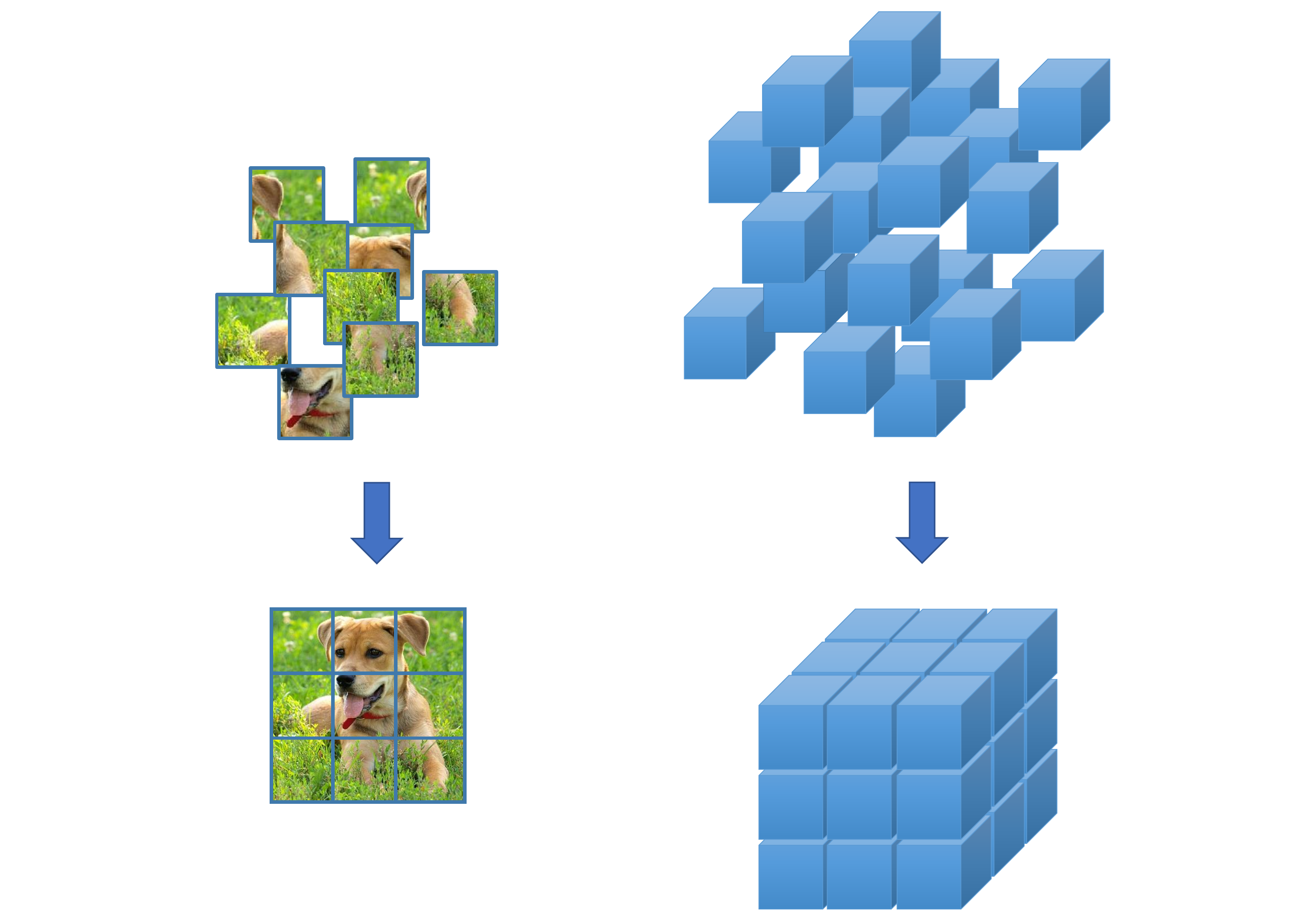}
\caption{
We study the problem of solving jigsaw puzzles for visual recognition. Compared to the previous work~\cite{noroozi2016unsupervised} which worked on $3\times3$ puzzles and $1\rm{,}000$ fixed configurations (left), we can solve this task in a generalized setting like 3D puzzles (right).
}
\label{Fig:JigsawPuzzles}
\end{figure}

Deep learning especially convolutional neural networks has been boosting the performance of a wide range of applications in computer vision~\cite{lecun2015deep}. These statistics-based approaches build hierarchical structures which contain a large number of neurons, so that visual knowledge is learned by fitting labeled training data~\cite{krizhevsky2012imagenet}. However, annotating a large-scale dataset is often difficult and expensive. Therefore, weakly supervised or unsupervised learning has attracted a lot of research attentions~\cite{weinberger2006unsupervised}\cite{le2013building}. These approaches are often built on some naturally existing constraints such as temporal consistency~\cite{wang2015unsupervised}, spatial relationship~\cite{doersch2015unsupervised} and sum-up equations~\cite{noroozi2017representation}. Such information, though being weak, constructs loss functions without requiring annotations, and networks pre-trained in this way can either be used for weak visual feature extraction~\cite{wang2015unsupervised} or fine-tuned in a standalone supervised learning process towards better recognition performance~\cite{erhan2010does}.

In this work, we focus on a specific way of exploiting spatial relationship, which is to solve {\em jigsaw puzzles} on unlabeled image data~\cite{noroozi2016unsupervised}\cite{noroozi2018boosting}. These approaches work by cutting an image into a grid, say, $3\times3$, of patches and then disordering them as training data, with the goal set to recover its correct spatial configuration. Examples are shown in Figure~\ref{Fig:JigsawPuzzles}. Thus, in order to achieve this goal, the network should have the ability to capture some semantic information, {\em e.g.}, learning the concept of {\em car} and {\em ground}, though not labeled, and knowing that {\em car} always appears above {\em ground}. Technically, these approaches simply assigned each configuration a unique ID, so that puzzle recognition turns into a {\em plain} classification problem. We point out two major drawbacks of this strategy. First, by plain classification, we assume that all configurations have the same similarity with each other, but this is often not the case, {\em e.g.}, two $3\times3$ configurations with only two patches swapped are often semantically closer than other two with no patches placed at the same position. Ignoring such information can bring in difficulties to representation learning. Second, the number of parameters required for plain classification increases linearly with the number of configurations, so that it is very difficult to deal with all possible configurations due to the risk of over-fitting. For example, there are ${9!}={362\rm{,}880}$ possible configurations for a $3\times3$ puzzle, but the original approach~\cite{noroozi2016unsupervised} reached the best performance at $1\rm{,}000$ and observed over-fitting when this number continues growing. Both of these drawbacks limit us from generalizing this approach to more complex puzzles\footnote{It was widely believed that more powerful features can be learned in more difficult vision tasks~\cite{deng2010does}, so we expect the ability of unsupervised learning to grow with the complexity of puzzles.} like 3D puzzles\footnote{This is especially useful for some areas such as medical imaging analysis, in which 3D networks~\cite{cicek20163d}\cite{milletari2016v} cannot easily get pre-trained weights as in 2D scenarios, yet a reasonable initialization helps a lot in training stability and testing performance.}. An empirical study of this topic can be found in Section~\ref{Experiments:TransferLearning}.

In this paper, we extend the ability of such approaches by allowing it to solve arbitrary jigsaw puzzles, {\em i.e.}, the puzzles are not constrained by a pre-defined set of configurations. Our major contribution is to provide a principle for unsupervised learning, that learning to recognize weak visual cues and then composing them into a complex scene is often easier and thus better in transfer. Thus, we solve jigsaw puzzles (i) in an iterative manner and (ii) using weak spatial cues, instead of determining the correct configuration all at once. To this end, we formulate puzzle recognition into an optimization problem which involves a set of unary and binary terms, with each unary term indicating whether a specified patch is located at a specified position, and each binary term measuring whether two patches should have a specified relative position. These terms are determined by a deep network backbone so that the entire system can be trained in an end-to-end manner. In both training and testing, we allow the first trial not to find the correct configuration, in which case we iterate using the configuration adjusted according to prediction until convergence. Both the above techniques, {\em a.k.a.}, network heads, are used to improve the quality of pre-training. They do not apply to nor introduce additional computational costs to the transfer learning stage.

We evaluate our approach in both puzzle recognition and transfer learning. The puzzle solver is trained on the ILSVRC2012 training set~\cite{russakovsky2015imagenet} tested on the validation set, both of which do not contain class labels. Our approach solves arbitrary jigsaw puzzles with reasonable accuracy, while the prior approaches can only work on a limited set of puzzle. Then, we transfer the pre-trained model to extract features in small-scale datasets for image classification~\cite{griffin2007caltech}, as well as to be fine-tuned in the PascalVOC 2007 dataset~\cite{everingham2010pascal} for image classification and object detection. Either learning from more complex puzzles or achieving a higher accuracy in puzzle recognition boosts transfer learning performance, which verifies our motivation. Finally, we apply our approach initialize a 3D network with unlabeled medical data, and verify its effectiveness in segmenting an abdominal organ from CT scans.

The remainder of this paper is organized as follows. Section~\ref{RelatedWork} briefly reviews related work, and Section~\ref{Approach} describes the proposed approach. After experiments are shown in Section~\ref{Experiments}, we draw our conclusions in Section~\ref{Conclusions}.

\section{Related Work}
\label{RelatedWork}

Deep neural networks have been playing an important role in modern computer vision systems. With the availability of large-scale datasets~\cite{deng2009imagenet} and powerful computational device such as GPUs, researchers have designed network structures with tens~\cite{krizhevsky2012imagenet}\cite{simonyan2015very}\cite{szegedy2015going} or hundreds~\cite{he2016deep}\cite{huang2017densely} of layers towards better recognition performance. Also, the pre-trained networks in ImageNet were transferred to other recognition tasks by either extracting visual features directly~\cite{donahue2014decaf}\cite{girshick2014rich}\cite{razavian2014cnn} or being fine-tuned on a new loss function~\cite{long2015fully}\cite{ren2015faster}. Despite their effectiveness, these networks still strongly rely on labeled image data, but in some areas such as medical imaging, data collection and annotation can be expensive, time-consuming, or requiring expertise. Thus, there has been efforts to design unsupervised~\cite{weinberger2006unsupervised}\cite{le2013building} or weakly supervised~\cite{joulin2016learning} approaches which learned visual knowledge from unlabeled data, or semi-supervised learning algorithms~\cite{papandreou2015weakly}\cite{qiao2018deep} which were aimed at combining a limited amount of labeled data and a large corpus of unlabeled data towards better performance. It has been verified that unsupervised pre-training helps supervised learning especially deep learning~\cite{erhan2010does}.

The key factor to learning from unlabeled data is to establish some kind of {\em prior}, or some weak constraints that naturally exist, {\em i.e.}, no annotations are required. Such prior can be either (1) embedded into the network architecture or (2) encoded as a weak supervision to optimize the network. For the first type, researchers designed clustering-based approaches to optimize visual representation so as to be beneficial to clustering~\cite{yang2016joint}\cite{caron2018deep}, as well as generator-based approaches which assumed that all images can be represented in a low-level space and trained encoders and/or decoders to recover the image and/or representation~\cite{radford2016unsupervised}\cite{zhu2017unpaired}. Network architectures of these approaches are often largely modified, {\em e.g.}, with a set of clustering layers or encoder-decoder modules.

This paper mainly considers the second type which, in comparison to the type, is much easier in algorithmic design. Typical examples include temporal consistency which assumes that neighboring video frames contain similar visual contents~\cite{wang2015unsupervised}, spatial relationship between some pairs of unlabeled patches~\cite{doersch2015unsupervised}, learning an additive function on different regions as well as the entire image~\cite{noroozi2017representation}, {\em etc}. Among these priors, spatial contexts are widely believed to contain rich information which a vision system should be able to capture. Going one step beyond modeling patch relationship~\cite{doersch2015unsupervised}, researchers designed so-called jigsaw puzzles~\cite{noroozi2016unsupervised}\cite{noroozi2018boosting} which are more complex so that the networks are better trained in learning to solve them. Consequently, such networks perform better in transfer learning.

Researchers believed that learning from these weakly-supervised cues can help visual recognition, because many problems are indeed built on understanding and integrating this type of information. Regarding spatial contexts, a wide range of recognition tasks can benefit from understanding the relative position of two (or more) patches, such as image classification~\cite{aghajanian2009patch}, semantic segmentation~\cite{rousseau2011supervised} and parsing~\cite{zhang2018deepvoting}, {\em etc.}

\section{Our Approach}
\label{Approach}

\subsection{Problem and Baseline Solution}
\label{Approach:Motivation}

The problem of puzzle recognition assumes that an image is partitioned into a grid ({\em e.g.}, $3\times3$) of patches and then disordered, and the task is to recover the original configuration ({\em i.e.}, patches are ordered in the natural form). To accomplish this task, the network needs to understand what a patch contains as well as how two or more patches are related to each other ({\em e.g.}, in a {\em car} image, a {\em wheel} is often located to the top of the {\em ground}). Therefore, we expect this task to teach a network both intra-patch and inter-patch information, which we formulate as unary terms and binary terms, respectively.

We first define the terminologies used in this paper. Let $\mathbf{I}$ be an image, which is partitioned into $W\times H$ patches. Each patch, denoted $\mathbf{i}_{x,y}$ (${0}\leqslant{x}<{W}$, ${0}\leqslant{y}<{H}$), is assigned a unique ID ${a_{x,y}}\in{\left\{0,1,\ldots,WH-1\right\}}$ according to its original position, {\em e.g.}, the row-major policy gives ${a_{x,y}}={x+yW}$. After that, all patches are randomly disordered, and we use $c_{x,y}^\star$ to denote the ID owned by the patch that currently occupies the $\left(x,y\right)$ position. All $c_{x,y}^\star$ values compose a configuration, denoted as ${\mathbf{c}^\star}={\left(c_{x,y}^\star\right)_{x=0,y=0}^{W,H}}$. There are in total $\left(WH\right)!$ different configurations, composing the configuration set $\mathcal{C}$ that ${\left|\mathcal{C}\right|}={\left(WH\right)!}$.

Our goal is to predict the correct configuration ${\mathbf{c}^\star}\in{\mathcal{C}}$. For this purpose, a network structure with two parts was constructed~\cite{noroozi2016unsupervised}. The network {\em backbone} ${\mathbb{M}^\mathrm{B}}:{\mathbf{f}_{x,y}}={\mathbf{f}\!\left(\mathbf{i}_{x,y};\boldsymbol{\theta}^\mathrm{B}\right)}$ is built upon each individual patch, and outputs a set of features for the network {\em head} ${\mathbb{M}^\mathrm{H}}:{\mathbf{c}}={\mathbf{g}\!\left(\mathbf{F};\boldsymbol{\theta}^\mathrm{H}\right)}$ to produce the final output ${\mathbf{c}}={\left(c_{x,y}\right)_{x=0,y=0}^{W,H}}$, where ${\mathbf{F}}={\left(\mathbf{f}_{x,y}\right)_{x=0,y=0}^{W,H}}$ is the ordered concatenation of patch features. In practice, $\mathbf{f}\!\left(\cdot;\boldsymbol{\theta}^\mathrm{B}\right)$ is often borrowed from existing network architectures~\cite{krizhevsky2012imagenet}\cite{simonyan2015very}\cite{he2016deep}, while $\mathbf{g}\!\left(\cdot;\boldsymbol{\theta}^\mathrm{H}\right)$ is often more interesting to investigate.

In the prior work~\cite{noroozi2016unsupervised}\cite{noroozi2018boosting}, the network head worked by constraining the number of possible configurations, say ${K}={1\rm{,}000}$ out of $9!$, which are randomly sampled from $\mathcal{C}$ using a greedy algorithm to guarantee the Hamming distance between any two configurations is sufficiently large. Then, $\mathbf{f}\!\left(\cdot;\boldsymbol{\theta}^\mathrm{H}\right)$ was designed to be a $K$-way classifier, implemented as a fully-connected layer. The purpose of this design was mainly to control the number of parameters of the classifier (proportional to $K$) so as to prevent over-fitting\footnote{\cite{noroozi2016unsupervised} observed that setting a larger $K$ leads to performance drop in transfer experiments, and explained it as the network gets confused by very similar jigsaw puzzles. However, as shown in experiments (see Section~\ref{Experiments:TransferLearning}), our approach works well in the entire puzzle set $\mathcal{C}$, {\em i.e.}, ${K}={9!}={362\rm{,}880}$, which implies that the performance drop may due the large number of parameters.}, but we argue that it largely limits the model from being applied more complex scenarios like 3D puzzles, while it was believed that learning from a harder task can lead to a stronger ability~\cite{deng2010does}. This motivates us to propose a new approach in which the number of configurations can be arbitrarily large while the number of parameters remains unchanged. We will see later that the essence behind this motivation is to use weak cues with an iterative algorithm towards a more compact representation and a safer learning process.

\subsection{Solving Jigsaw Puzzles with Weak Cues}
\label{Approach:Solution}

\begin{figure*}
\centering
\includegraphics[width=16.5cm]{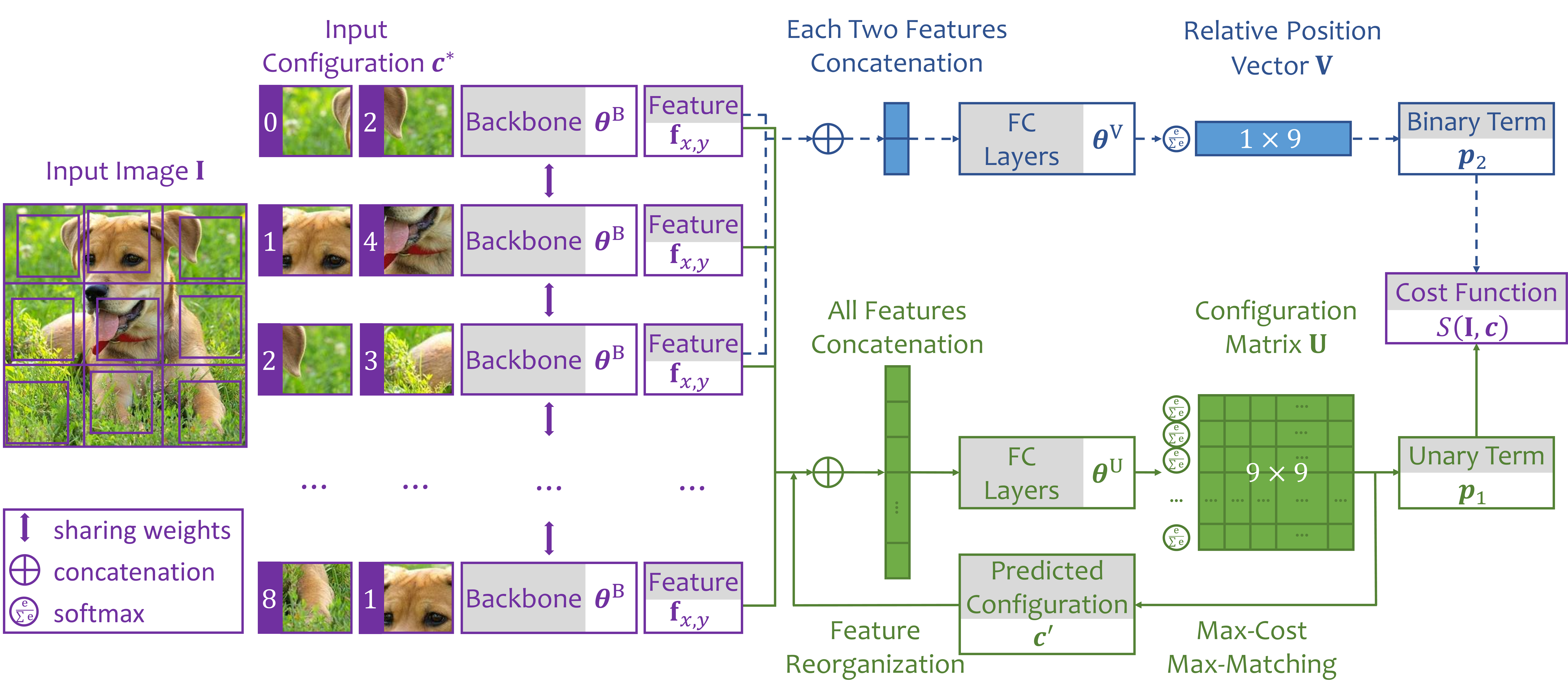}
\caption{
The overall structure (best viewed in color). Each training image (without semantic annotations) is randomly cropped, disordered and fed into puzzle recognition network. Two types of loss terms (unary and binary) are computed and summed into the final cost function $S\!\left(\mathbf{I},\mathbf{c}\right)$. The training process continues until the puzzle is completely correct or a maximal number of rounds is achieved.
}
\label{Fig:Structure}
\end{figure*}

We design a network head to learn {\em weak spatial constraints}. By ``weak'' we are comparing this strategy with the aforementioned $K$-way classifier that predicts the configuration of the entire puzzle all at once. Instead, we consider an indirect cost function $S\!\left(\mathbf{I},\mathbf{c}\right)$ which outputs a cost that patch $\mathbf{i}_{x,y}$ or equivalently feature $\mathbf{f}_{x,y}$ is located at position $c_{x,y}$, and thus the most probable configuration is determined by $\arg\max_\mathbf{c}\left\{S\!\left(\mathbf{I},\mathbf{c}\right)\right\}$. $S\!\left(\mathbf{I},\mathbf{c}\right)$ is composed of two parts, namely, unary terms and binary terms. Each {\em unary term} provides cues for the {\em absolute} position of a patch, and each {\em binary term} provides cues for the {\em relative} position of two patches. Mathematically,
\begin{eqnarray}
\nonumber
{S\!\left(\mathbf{I},\mathbf{c}\right)}\equiv{S\!\left(\mathbf{F},\mathbf{c}\right)}={{\sum_{\left(x,y\right)}}p_1\!\left(\mathbf{f}_{x,y},c_{x,y}\mid\mathbf{F}\right)+}\quad\quad\\
\label{Eqn:ScoreFunction}
{\quad\quad{\sum_{\left(x_1,y_1\right)\neq\left(x_2,y_2\right)}}p_2\!\left(\mathbf{f}_{x_1,y_1},\mathbf{f}_{x_2,y_2},c_{x_1,y_1},c_{x_2,y_2}\right)}.
\end{eqnarray}
Here, $p_1\!\left(\mathbf{f}_{x,y},c_{x,y}\mid\mathbf{F}\right)$ is a unary term which measures how likely that patch $f_{x,y}$ is located at position $c_{x,y}$, and $p_2\!\left(\mathbf{f}_{x_1,y_1},\mathbf{f}_{x_2,y_2},c_{x_1,y_1},c_{x_2,y_2}\right)$ is a binary term measures how likely that patches $\mathbf{f}_{x_1,y_1}$ and $\mathbf{f}_{x_2,y_2}$ have the spatial relationship indicated by $c_{x_1,y_1}$ and $c_{x_2,y_2}$. Each unary term is computed based on $\mathbf{F}$, the overall variable containing feature vectors of all patches, because the position of each patch $\mathbf{f}_{x,y}$ depends on the visual messages delivered by other patches. The binary terms, on the other hand, do not have such a dependency.

In practice, the unary terms are formulated in a matrix $\mathbf{U}$ with $WH\times WH$ elements, each of which, $\left\llbracket\mathbf{U}\right\rrbracket_{a,c}$, indicates the cost obtained by putting the specified patch with ID $a$ at a specified position with ID $c$. This is implemented by a fully-connected layer between $\mathbf{F}$ and these $\left(WH\right)^2$ elements, parameterized by $\boldsymbol{\theta}^\mathrm{U}$. We perform the softmax function over all elements in each row, so that the scores corresponding to each patch sum to $1$\footnote{Ideally, the elements in each column should also sum to $1$, but it is mathematically intractable if we hope to keep the ratio between all elements. There are two arguments. First, after normalizing scores in each row, we find that there often exists one major elements in each column, and the sum of each column is close to $1$. Second, we add an additional $\ell_1$ loss term between the sum of each column and $1$, but only observe to minor changes in either puzzle recognition accuracy or transfer learning performance.}. Then, each unary term is the log-likelihood of the score at a specified position:
\begin{equation}
\label{Eqn:UnaryTerm}
{p_1\!\left(\mathbf{f}_{x,y},c_{x,y},\mathbf{F}\right)}={-\ln\left\llbracket\mathbf{U}\!\left(\mathbf{F};\boldsymbol{\theta}^\mathrm{U}\right)\right\rrbracket_{a_{x,y},c_{x,y}}}.
\end{equation}

For each binary term involving $\mathbf{f}_{x_1,y_1}$ and $\mathbf{f}_{x_2,y_2}$, we build another mapping from these two vectors to a $9$-dimensional vector, with each index indicating the probability that the spatial relationship of $\mathbf{f}_{x_1,y_1}$ and $\mathbf{f}_{x_2,y_2}$ belongs to one of the $9$ possibilities, namely, the first patch is located to the top, bottom, left, right, top-left, top-right, bottom-left, bottom-right of the second patch or none of the above happens. Similarly, this is implemented using another fully-connected layer between $\mathbf{f}_{x_2,y_2}\oplus\mathbf{f}_{x_2,y_2}$ ($\oplus$ denotes concatenation) and a $9$-dimensional vector parameterized by $\boldsymbol{\theta}^\mathrm{V}$ followed by a softmax activation over these $9$ numbers. We denote ${r_{x_1,y_1,x_2,y_2}}\doteq{r\!\left(c_{x_1,y_1},c_{x_2,y_2}\right)}\in{\left\{0,1,\ldots,8\right\}}$ as the relative position type between $\mathbf{f}_{x_1,y_1}$ and $\mathbf{f}_{x_2,y_2}$, so that we can write the binary term as:
\begin{eqnarray}
\nonumber
{p_2\!\left(\mathbf{f}_{x_1,y_1},\mathbf{f}_{x_2,y_2},c_{x_1,y_1},c_{x_2,y_2}\right)}=\quad\quad\quad\quad\quad\quad\\
\label{Eqn:BinaryTerm}
\quad\quad{-\ln\left\llbracket\mathbf{V}\!\left(\mathbf{f}_{x_1,y_1},\mathbf{f}_{x_2,y_2};\boldsymbol{\theta}^\mathrm{V}\right)\right\rrbracket_{r_{x_1,y_1,x_2,y_2}}}.
\end{eqnarray}

Compared to a plain classifier assigning a class index to each puzzle, the amount of parameters required by our approach is reduced. Take a $3\times3$ puzzle as an example, and we assume that $F$ contains $D$ elements. On the one hand, the $K$-way classifier requires $KD$ parameters (a typical setting~\cite{noroozi2016unsupervised} is ${K}={1\rm{,}000}$) which grows linearly with $K$. On the other hand, our approach requires $ \left( WH \right)^2D$ parameters for the unary terms, and $9D$ parameters for the binary terms. The total number of parameters, $\left(W^2H^2+9\right)D$ ({\em e.g.}, $90D$ for a $3\times3$ puzzle), is largely reduced and does not increase with $K$. Consequently, our approach is easier to be applied to the scenario with a larger set of ({\em e.g.}, all $9!$ possible) configurations. This advantage is verified in experiments.

Last but not least, there are many other ways of using weak spatial constraints to formulate $S\!\left(\mathbf{I},\mathbf{c}\right)$ -- we just provide a practical example.

\subsection{Optimization: Iterative Reorganization}
\label{Approach:Optimization}

We aim at optimizing $S\!\left(\mathbf{F},\mathbf{c}\right)$ with respect to network parameters $\boldsymbol{\theta}^\mathrm{U}$, $\boldsymbol{\theta}^\mathrm{V}$ and configuration $\mathbf{c}$. However, note that $\mathbf{c}$ is a discrete variable which cannot be optimized by gradient descent. So we apply different strategies in training and testing.

In the training stage, we know the ground-truth configuration $\mathbf{c}^\star$, so the optimization becomes:
\begin{equation}
\label{Eqn:OptimizationTraining}
\arg\min_{\boldsymbol{\theta}^\mathrm{U},\boldsymbol{\theta}^\mathrm{V}}S\!\left(\mathbf{F},\mathbf{c}^\star\right).
\end{equation}
This is implemented by setting the supervision signal accordingly, {\em i.e.}, the correct cells are filled up with $1$ while others with $0$, and using stochastic gradient descent. Note that each unary term depends on the order of input patches\footnote{We fully-connect $\mathbf{F}$ to the $WH\times WH$ matrix, which is an asymmetric function and thus makes the output sensitive to the order of input. We can also design a symmetric function to deal with this issue, {\em e.g.}, each patch $\mathbf{f}_{x,y}$ is concatenated with the average-pooled vector of other patches to form the input, but this often causes information loss and leads to lower accuracy in both puzzle recognition and transfer learning tasks.}. To sample more training data as well as adjust data distribution (explained later), we introduce iteration to the training stage. Denote the input configuration as ${\mathbf{c}^{\left(0\right)}}={\mathbf{c}^\star}$, and the corresponding feature as $\mathbf{F}^{\left(0\right)}$. In each iteration, with fixed $\boldsymbol{\theta}^\mathrm{U}$ and $\boldsymbol{\theta}^\mathrm{V}$, we maximize $S\!\left(\mathbf{F},\mathbf{c}\right)$ with respect to $\mathbf{c}$:
\begin{equation}
\label{Eqn:OptimizationTesting}
{\mathbf{c}'}={\arg\min_{\mathbf{c}}S\!\left(\mathbf{F},\mathbf{c}^{\left(0\right)}\right)},
\end{equation}
and use $\mathbf{c}'$ to find the next input $\mathbf{c}^{\left(1\right)}$, so that applying $\mathbf{c}'$ to $\mathbf{c}^{\left(1\right)}$ obtains $\mathbf{c}^{\left(0\right)}$, {\em e.g.}, if $\mathbf{c}'$ is perfect, then $\mathbf{c}^{\left(1\right)}$ corresponds to the original configuration that every patch is placed at the correct position. This process continues until convergence or a maximal number of iterations is reached. The losses with respect to $\boldsymbol{\theta}^\mathrm{U}$ and $\boldsymbol{\theta}^\mathrm{V}$ are accumulated, averaged, and back-propagated to update these two parameters. The same strategy, iteration, is used at the testing stage to solve jigsaw puzzles, with the only difference that no gradient back-propagation is required.

It remains a problem to solve Eqn~\eqref{Eqn:OptimizationTesting}. This is a combinatoric optimization problem, as $\mathbf{c}$ can only take $\left(WH\right)!$ discrete values which indicate the entries in $\mathbf{U}$ and $\mathbf{V}$ that are summed up. There is obviously no closed form solutions to maximize $S\!\left(\mathbf{F},\mathbf{c}\right)$, yet enumerating all $\left(WH\right)!$ possibilities is computationally intractable especially when the puzzle size becomes large. A possible solution lies in approximation, which first switches off all binary terms, so that the optimization becomes choosing $WH$ entries from a $WH\times WH$ matrix with a maximal sum, but no two entries can appear in the same row or column (this is a max-cost-max-matching problem, and the best solution $\tilde{\mathbf{c}}$ can be found using the Hungarian algorithm); then enumerates all possibilities within a limited Hamming distance from $\tilde{\mathbf{c}}$ and chooses the one with the best overall cost $S\!\left(\mathbf{F},\mathbf{c}\right)$.

Finally, we discuss strategy of introducing iteration to solve this problem. Mathematically, Eqn~\eqref{Eqn:OptimizationTesting} is a fixed-point model~\cite{li2013fixed}, {\em i.e.}, the output variable $\mathbf{c}$ also impacts $\mathbf{F}$ and thus $S\!\left(\mathbf{F},\mathbf{c}\right)$, so iteration is considered a regular way of optimizing it. However, the roles played by iteration are different in training and testing. In the {\bf training stage}, after each iteration, we shall expect the configuration to be adjusted closer to the ground-truth. Therefore, if we take the input configuration fed into each round as an individual case, then the distribution of input data is changed by iteration, and the cases that are more similar to the ground-truth are more likely to be sampled. Therefore, in the {\bf testing stage}, we can expect the iteration to improve puzzle recognition accuracy, because as the iteration continues, the input puzzle gets closer to the ground-truth by statistics, and our model sees more training data in this scenario and is stronger. We show a typical example in Figure~\ref{Fig:PuzzleRecognition}, in which we can observe how iteration gradually predicts the correct configuration.

\section{Experiments}
\label{Experiments}

\subsection{Jigsaw Puzzle Recognition}
\label{Experiments:JigsawPuzzles}

\newcommand{\colwidthA}{1.1cm}
\newcommand{\colwidthB}{1.3cm}
\begin{table*}[!btp]
\centering{
\setlength{\tabcolsep}{0.16cm}
\begin{tabular}{|L{\colwidthA}|L{\colwidthA}|l||C{\colwidthA}|C{\colwidthA}|C{\colwidthA}|C{\colwidthA}||R{\colwidthB}|R{\colwidthB}||R{\colwidthB}|R{\colwidthB}|}
\hline
{}  & \multicolumn{2}{c||}{Setting} & \multicolumn{4}{c||}{Pre-training Options} & \multicolumn{2}{c||}{Puzzle Recognition} & \multicolumn{2}{c|}{PascalVOC 2007} \\
\hline
ID  & Size       & Backbone & Label      & Unary      & Binary     & Mirror     & Correct & $D\leqslant2$     & Classifi. & Detec. \\
\hline\hline
(a) & $3\times3$ & AlexNet  & \checkmark &            &            &            &     $-$ &           $-$ & $78.2$ & $56.8$ \\
\hline
(b) & $3\times3$ & AlexNet  &            &            &            &            &     $-$ &           $-$ & $53.3$ & $43.3$ \\
\hline
(c) & $3\times3$ & AlexNet  &            & \checkmark &            &            &  $32.2$ &        $48.2$ & $66.6$ & $51.8$ \\
\hline
(d) & $3\times3$ & AlexNet  &            & \checkmark &            & \checkmark &  $ 3.4$ &        $15.4$ & $68.1$ & $51.9$ \\
\hline
(e) & $3\times3$ & AlexNet  &            & \checkmark & \checkmark & \checkmark &  $ 3.8$ &        $17.1$ & $68.3$ & $52.5$ \\
\hline\hline
(f) & $2\times2$ & AlexNet  &            & \checkmark & \checkmark & \checkmark &  $74.5$ &        $90.5$ & $64.2$ & $49.1$ \\
\hline\hline
(g) & $3\times3$ & ResNet18 & \checkmark &            &            &            &     $-$ &           $-$ & $84.5$ & $68.3$ \\
\hline
(h) & $3\times3$ & ResNet18 &            &            &            &            &     $-$ &           $-$ & $41.3$ & $24.8$ \\
\hline
(i) & $3\times3$ & ResNet18 &            & \checkmark &            &            &  $44.7$ &        $61.5$ & $72.5$ & $58.7$ \\
\hline
(j) & $3\times3$ & ResNet18 &            & \checkmark &            & \checkmark &  $ 5.2$ &        $20.4$ & $72.9$ & $58.7$ \\
\hline
(k) & $3\times3$ & ResNet18 &            & \checkmark & \checkmark & \checkmark &  $ 5.5$ &        $21.0$ & $74.7$ & $58.8$ \\
\hline\hline
(l) & $3\times3$ & ResNet50 & \checkmark &            &            &            &     $-$ &           $-$ & $86.4$ & $70.2$ \\
\hline
(m) & $3\times3$ & ResNet50 &            &            &            &            &     $-$ &           $-$ & $46.8$& $23.5$ \\
\hline
(n) & $3\times3$ & ResNet50 &            & \checkmark &            &            &  $47.3$ &        $63.6$ & $72.4$ & $55.2$ \\
\hline
(o) & $3\times3$ & ResNet50 &            & \checkmark &            & \checkmark &  $ 4.9$ &        $20.4$ & $73.1$ & $55.5$ \\
\hline
(p) & $3\times3$ & ResNet50 &            & \checkmark & \checkmark & \checkmark &  $ 5.2$ &        $20.8$ & $75.3$ & $56.2$ \\
\hline
\multicolumn{11}{c}{Competitors with Different Backbones, Pre-training Cues and Settings} \\
\hline
Ref.& Year       & Backbone & \multicolumn{6}{c||}{Description of unsupervised training} & Classifi. &  Detec. \\
\hline
\cite{doersch2015unsupervised}
    & 2015       & AlexNet  & \multicolumn{6}{l||}{{\em Determining the relative spatial position of two patches}} & $65.3$ & $51.1$ \\
\hline
\cite{wang2015unsupervised}
    & 2015       & AlexNet  & \multicolumn{6}{l||}{{\em Unsupervised tracking in videos}} & $63.1$ & $47.2$ \\
\hline
\cite{noroozi2016unsupervised}
    & 2016       & AlexNet  & \multicolumn{6}{l||}{\textcolor{green}{\em $3\times3$ jigsaw puzzles with a $1\rm{,}000$-way plain classifier}} & $67.7$ & $53.2$ \\
\hline
\cite{larsson2017colorization}
    & 2017       & ResNet152  & \multicolumn{6}{l||}{{\em Predicting color from gray-scale intensity}} & $77.3$ & $-$ \\
\hline
\cite{noroozi2017representation}
    & 2017       & AlexNet  & \multicolumn{6}{l||}{{\em Counting visual primitives in subregions}} & $67.7$ & $51.4$ \\
\hline
\cite{caron2018deep}
    & 2018       & AlexNet  & \multicolumn{6}{l||}{{\em Classifying after clustering iteratively}} & $73.7$ & $55.4$ \\
\hline
\cite{gidaris2018unsupervised}
    & 2018       & AlexNet  & \multicolumn{6}{l||}{{\em Predicting 2D image rotations}} & $73.0$ & $54.4$ \\
\hline
\cite{mundhenk2018improvements}
    & 2018       & AlexNet  & \multicolumn{6}{l||}{{\em \cite{doersch2015unsupervised} with enhancement techniques}} & $69.6$ & $55.8$ \\
\hline
\cite{noroozi2018boosting}
    & 2018       & VGGNet16 & \multicolumn{6}{l||}{\textcolor{green}{\em \cite{noroozi2016unsupervised} with knowledge distillation and noisy patches}} & $72.5$ & $56.5$ \\
\hline
\cite{ren2018cross}
    & 2018       & AlexNet  & \multicolumn{6}{l||}{{\em Predicting surface normal, depth, and instance contour}} & $68.0$ & $52.6$ \\
\hline
\end{tabular}}
\caption{Puzzle recognition and transfer learning accuracy ($\%$). In the pre-training options, ``labeled'' means to use the annotated ILSVRC2012 training set to pre-train a network. The instances without any \checkmark imply that PascalVOC 2007 tasks are trained from scratch. We also compare with prior approaches, some of which have different knowledge sources, network backbones and training strategies. We report the most powerful network backbone used in each paper. The works with puzzle recognition are highlighted in \textcolor{green}{green}.}
\label{Tab:2DResults}
\end{table*}

We follow~\cite{noroozi2016unsupervised} to train and evaluate puzzle recognition on the ILSVRC2012 dataset~\cite{russakovsky2015imagenet}, a subset of the ImageNet database~\cite{deng2009imagenet}. We train the model using all the $1.3\mathrm{M}$ training images and test it on the validation set with $50\mathrm{K}$ images, both of which do not contain class annotations.

In the training stage, we pre-process the images to prevent the model from being disturbed by pixel-level information. We first determine the size of puzzles, {\em e.g.}, $W\times H$, and then resize each input image into $85W\times85H$ and partition it evenly into a $W\times H$ grid. In each $85\times85$ image, we randomly crop a $64\times64$ subimage as the patch fed into the puzzle recognition network. To maximally reduce the possibility that low-level information is used, we further horizontally flip each input patch with a probability of $50\%$ and subtract mean value from each channel -- we do not perform other data augmentation techniques because they are less likely to appear in real data. In practice, flip augmentation brings consistent accuracy gain to transfer learning tasks though we observe significant accuracy drop in puzzle recognition (see Table~\ref{Tab:2DResults}).

\begin{figure*}
\centering
\includegraphics[width=16.5cm]{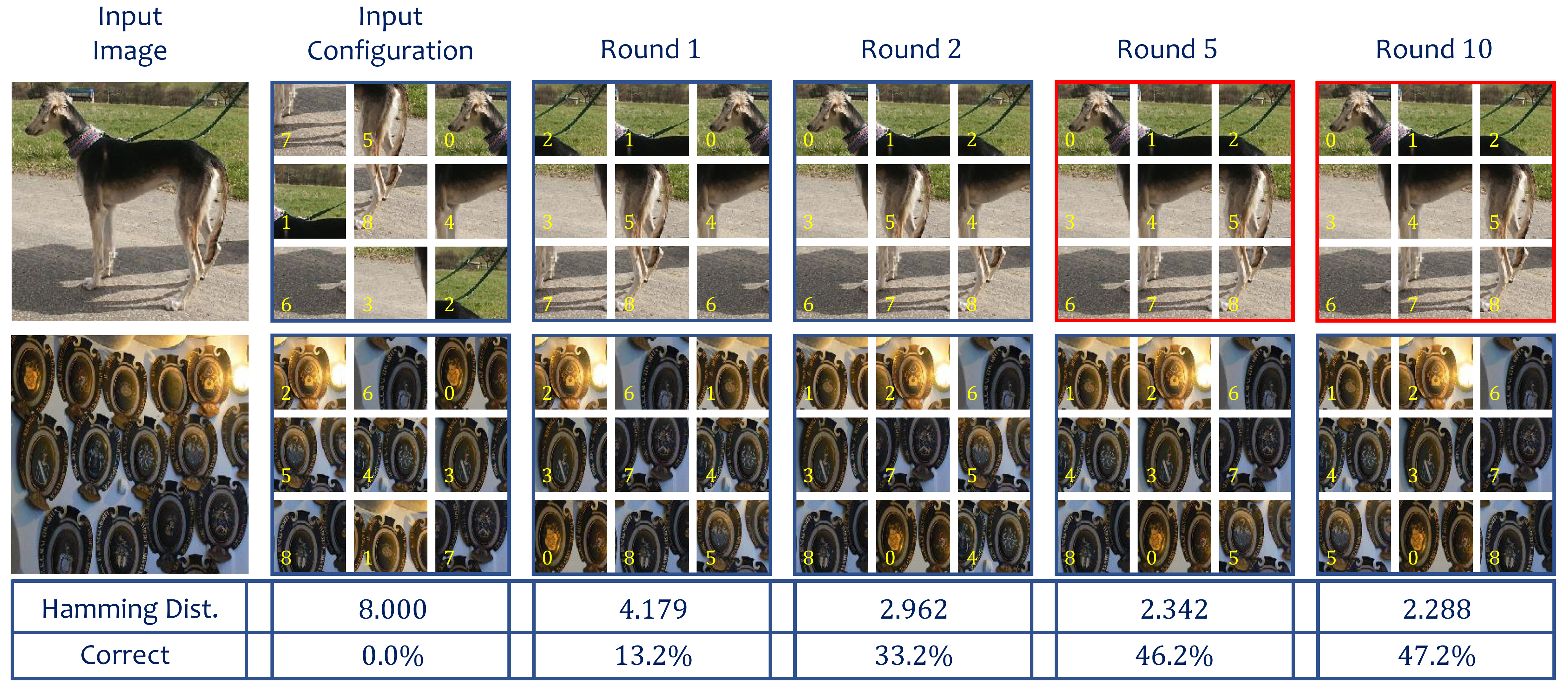}
\caption{Two examples of different difficulties in iterative puzzle recognition (best viewed in color). Each digit to the lower-left corner of each patch is the corresponding patch ID. For each round, we also report puzzle recognition statistics {\bf over the entire testing set}.
}
\vspace{-0.1 cm}
\label{Fig:PuzzleRecognition}
\end{figure*}

The backbone of our puzzle network is borrowed from two popular architectures, namely, an $8$-layer AlexNet~\cite{krizhevsky2012imagenet} and two deep ResNets~\cite{he2016deep} with $18$ and $50$ layers. We do not evaluate VGGNet~\cite{simonyan2015very} as in~\cite{larsson2017colorization}\cite{noroozi2018boosting} because it is more difficult to initialize and produces lower accuracy than ResNets. The outputs of the first layer with a spatial resolution of $1\times1$ ({\em i.e.}, {\em fc6} in AlexNet and {\em avg-pool} in ResNets) are fed into a $1\rm{,}024$-way fully-connected layer and the output is taken as $\mathbf{f}_{x,y}$, followed by our designed layers for extracting unary and binary terms for puzzle recognition. All these networks are trained from scratch. We use the SGD optimizer and a total of $250\mathrm{K}$ iterations (mini-batches) for AlexNet and $350\mathrm{K}$ for ResNets. Each batch contains $256$ puzzles. On four NVIDIA Titan-V100 GPUs, the training times on AlexNet, ResNet18 and ResNet50 are $10$, $20$ and $60$ hours, respectively.


In the testing stage, to reduce randomization factors, we switch off randomization in patch cropping and data augmentation, with each $64\times64$ patch cropped at the center of the $85\times85$ fields and not flipped. Results are summarized in Table~\ref{Tab:2DResults}. We first evaluate $3\times3$ puzzle recognition accuracy. For each image, there are ${9!}={362\rm{,}880}$ possible puzzles, so random guess gives a $0.0003\%$ accuracy. With only unary terms (Eqn~\ref{Eqn:OptimizationTesting} can be solved by the Hungarian algorithm), all network backbones achieve over $30\%$ accuracy without mirror augmentation, which shows that weak visual cues can be combined to infer global patch contexts.

On top of this baseline, we investigate the impact of other four options. {\bf First}, adding binary terms consistently improves puzzle recognition accuracy, arguably due to the additional contextual information, which is especially useful in determining the relative position of two neighboring patches. {\bf Second}, mirror augmentation reduces puzzle recognition accuracy dramatically in both training and testing, but as we will see later, this strategy improves the generalization ability of our pre-trained models to other recognition tasks. {\bf Third}, compared with $2\times2$ puzzles, $3\times3$ jigsaw puzzles are naturally more difficult to solve, but they also force the model to learn more visual knowledge and thus help transfer learning, as shown in our later discussions. {\bf Fourth}, the above phenomena remain the same as the network backbone becomes stronger, on which both puzzle recognition and transfer visual recognition becomes more accurate.

As a side comment, we point out that conventional puzzle recognition approaches with plain classification~\cite{noroozi2016unsupervised}\cite{noroozi2018boosting} often achieved higher puzzle recognition accuracy in a limited class set. With models trained with our approach (Line (e) in Table~\ref{Tab:2DResults}) we enumerate the $1\rm{,}000$ classes generated with algorithm provided by~\cite{noroozi2016unsupervised} and find the maximal $S\!\left(\mathbf{F},\mathbf{c}\right)$, so as to mimic the behavior of plain classification. Our models with AlexNet reports a $60.2\%$ puzzle recognition accuracy which is lower than $71\%$ reported in~\cite{noroozi2016unsupervised}. However, our approach enjoys better transfer ability, as we will see in later experiments. In addition, the performance of~\cite{noroozi2016unsupervised} degenerates with increased puzzle size, as the fraction of explored puzzles becomes smaller, yet the weakness of ignoring underlying relationship between different configurations becomes more significant and harmful. From this perspective, the advantage of solving arbitrary puzzles becomes clearer. The same phenomenon also happens in 3D puzzles (Section~\ref{Experiments:3DNets}).

Some statistics for our model with ResNet50 (Line (p) in Table~\ref{Tab:2DResults}) as well as two typical examples are shown in Figure~\ref{Fig:PuzzleRecognition} (one is difficult and not solved). We can observe how the disordered patches are reorganized with weak spatial cues throughout an iterative process. As an ablation study, we experiment with fewer numbers of maximal iterations, namely $1$, $5$ and $10$ instead of $20$, but achieve lower accuracies in both puzzle recognition and transfer learning tasks. This justifies our hypothesis that iteration, together with weak spatial cues, provides a mild way of unsupervised learning, which better fits state-of-the-art deep networks.

\subsection{Transfer Learning Performance}
\label{Experiments:TransferLearning}

Next, we investigate how well our models pre-trained on puzzle recognition transfer to other visual recognition tasks. Following the conventions~\cite{noroozi2018boosting}\cite{caron2018deep}, we evaluate classification and detection tasks on the PascalVOC 2007 dataset~\cite{everingham2010pascal}. All pre-trained networks undergo a standard fine-tuning flowchart, with a plain classifier and Fast-RCNN~\cite{girshick2015fast} being used as network heads, respectively. We do not lock any layers in our network, because this often leads to worse transfer performance as shown in prior approaches~\cite{noroozi2016unsupervised}\cite{caron2018deep}\cite{gidaris2018unsupervised}.

Results are summarized in Table~\ref{Tab:2DResults}. We can observe some interesting phenomena. {\bf First}, transfer recognition performance goes up with the power of network backbones, which shows the ability of our approach to tap the potential of deep networks. {\bf Second}, both unary and binary terms contribute to transfer accuracy and they are complementary. {\bf Third}, mirror augmentation harms puzzle recognition but improves transfer learning, because it alleviates the chance that deep networks borrow low-level pixel continuity in solving the jigsaw puzzles which falls into the category of over-fitting and helps transfer recognition very little.

Here is a side note. It was suggested in~\cite{noroozi2016unsupervised} that forcing the network to discriminate very similar puzzles ({\em e.g.}, only a pair of patches are reversed) often leads to accuracy drop because the model can focus too much on local patterns. In the context of using AlexNet to solve $3\times3$ puzzles, we study different numbers of configurations, {\em i.e.}, $1\%$ ($3\rm{,}629$), $10\%$ ($36\rm{,}288$) and all (${9!}={362\rm{,}880}$) possible puzzles. We find that our approach reports the best transfer accuracy at the last option, while using smaller numbers of configurations leads to slightly worse performance. Hence, we make the following conjecture: it is indeed the larger number of parameters in a plain classifier, rather than solving very similar puzzles, that causes transfer performance drop.

Last, we evaluate the quality of features extracted from the pre-trained models directly (the first fully-connected layer, without being fine-tuned). We apply a linear SVM with ${C}={10}$ to the Caltech256 dataset~\cite{griffin2007caltech} for generic object classification. Our $3\times3$ model based on AlexNet with unary terms, binary terms and mirror augmentation (Line (e) in Table~\ref{Tab:2DResults}) reports a $29.05\%$ accuracy, but our direct competitors~\cite{noroozi2016unsupervised} and~\cite{noroozi2018boosting} only reports $20.83\%$ and $23.07\%$, respectively, almost of the same quality as a randomly-initialized AlexNet ($18.73\%$).

\subsection{Generalization to 3D Networks}
\label{Experiments:3DNets}

Finally, we apply our model to a 3D visual recognition task, which lies in the area of medical imaging analysis, an important prerequisite for computer-assisted diagnosis (CAD). Most medical data are volumetric ({\em i.e.}, appearing in a 3D form), and researchers have proposed some 3D network architectures~\cite{cicek20163d}\cite{milletari2016v}. Compared to 2D networks~\cite{ronneberger2015u}\cite{yu2018recurrent}, 3D networks enjoy the benefit of seeing more contextual information, but still suffer the drawback of missing a pre-trained model. Due to the common situation that the amount of training data is limited, these 3D networks often have a relatively unstable training process and sometimes this downgrades their testing accuracy~\cite{xia2018bridging}.

Our approach provides a solution for initializing 3D networks with jigsaw puzzles. We investigate the NIH pancreas segmentation dataset~\cite{roth2015deeporgan}, which contains $82$ cases. We partition it into $4$ folds (around $20$ cases in each fold), use three of them to train a segmentation model and test it on the remaining one. To construct jigsaw puzzles, we either directly use the training samples in the NIH dataset, or refer to another public dataset named Medical Segmentation Decathlon (MSD)\footnote{{\tt http://medicaldecathlon.com/}} -- the {\em pancreas tumour} subset with $282$ training cases. For all the data used for jigsaw puzzles, we do not use any pixel-level annotations though they are provided. We randomly crop $120\times120\times120$ volumes within each case, and cut it evenly into two puzzle sizes, namely, $2\times2\times2$ pieces with a $48\times48\times48$ subvolume cropped within each cell, or $3\times3\times3$ pieces with a $32\times32\times32$ subvolume cropped within each cell. A typical example is shown in Figure~\ref{Fig:JigsawPuzzles}. We randomly disorder these patches using all $8!$ or $27!$ possible configurations, and the task is to recover the original configuration. We use VNet~\cite{milletari2016v} as the baseline (only the down-sampling layers are used in this stage), and compute the unary terms in an $8\times8$ or $27\times27$ matrix. We switch off the binary terms based on the consideration that one patch has $26$ neighbors in the 3D space which makes prediction over-complicated.

\renewcommand{\colwidthA}{0.84cm}
\renewcommand{\colwidthB}{1.08cm}
\newcommand{\colwidthC}{1.4cm}

\begin{table}[!btp]
\centering{
\setlength{\tabcolsep}{0.08cm}
\begin{tabular}{|C{\colwidthA}|C{\colwidthB}|C{\colwidthC}|C{\colwidthC}|C{\colwidthC}|C{\colwidthC}|}
\hline
\multirow{2}{*}{Data} & \multirow{2}{*}{Scratch} & \multicolumn{2}{c|}{Pre-trained on NIH} & \multicolumn{2}{c|}{Pre-trained on MSD} \\
\cline{3-6}
{}                    & {}                       & $2\times2\times2$  & $3\times3\times3$  & $2\times2\times2$  & $3\times3\times3$  \\
\hline\hline
$10\%$                & $65.52$                  & $69.36$            & $70.80$            & $68.44$            & $72.24$            \\
\hline
$20\%$                & $74.78$                  & $76.30$            & $76.50$            & $76.58$            & $77.80$            \\
\hline
$100\%$               & $80.96$                  & $79.88$            & $81.68$            & $81.48$            & $82.33$            \\
\hline
\end{tabular}}
\caption{Pancreas segmentation accuracy (DSC, $\%$) with different amounts of training data and different initialization techniques. In each group, the accuracy is averaged over $20$ testing cases.}
\label{Tab:3DResults}
\end{table}

Now we recover the complete VNet structure with randomly-initialized up-sampling layers and start training on the NIH training set ($62$ cases) as well as its subsets. Results are shown in Table~\ref{Tab:3DResults} revealing some useful knowledge. {\bf First}, pre-training on jigsaw-puzzles indeed helps segmentation especially in the scenarios of fewer training data. {\bf Second}, visual knowledge learned in this manner can transfer across different datasets regardless of the different distributions in intensity (caused by the scanning device). {\bf Third}, constructing larger and thus more difficult puzzles improves the basic ability of networks. This the value of our research -- note that it is unlikely for the baseline approach to sufficiently explore the space of $3\times3\times3$ puzzles, which has ${27!}\approx{1.1\times10^{28}}$ different configurations.

\section{Conclusions}
\label{Conclusions}

This work generalizes the framework of jigsaw puzzle recognition which was previously studied in a constrained case. To this end, we change the network head from a plain $K$-way classifier to a combinatoric optimization problem which uses both unary and binary weak spatial cues. This strategy reduces the number of learnable parameters in the model, and thus alleviates the risk of over-fitting. The increased flexibility of pre-training allows us to apply our approach to a wide range of transfer learning tasks, including directly using it for feature extraction, and generalizing it to the 3D scenarios to provide an initialization for other tasks, {\em e.g.}, medical imaging segmentation.

Our study reveals the ease and benefits of learning to recognize weak visual cues in unsupervised learning, in which the key problem often lies in finding a compact way of representing knowledge, {\em e.g.}, decomposing the entire puzzle into unary and binary terms. We point out that the exploration of unsupervised learning is still far from the end. In the future, we will also apply our method to less structured data such as graphs~\cite{kipf2017semi} and more structured data such as videos~\cite{karpathy2014large}, and explore its ability of learning visual knowledge in an unsupervised manner.

{\small
\bibliographystyle{ieee}
\bibliography{egbib}
}

\end{document}